\documentclass{article}
\usepackage{ijcai22}

\usepackage{times}

\usepackage{soul}
\usepackage{url}
\usepackage[hidelinks]{hyperref}
\usepackage[utf8]{inputenc}
\usepackage[small]{caption}
\usepackage{graphicx}
\usepackage{amsmath}
\usepackage{booktabs}
\usepackage{color}
\usepackage{multirow}
\title{
Knowledge Graph Reasoning with Logics and Embeddings: Survey and Perspective
}

\author{
Wen Zhang$^1$\footnote{Corresponding Authors.} \and
Jiaoyan Chen$^2$\and
Juan Li$^4$ \and
Zezhong Xu$^{4}$ \and 
Jeff Z. Pan$^{3}$ \And
Huajun Chen$^{4,5,6*}$
\\
\affiliations
$^1$School of Software Technology, Zhejiang University \\
$^2$Department of Computer Science, University of Oxford \\
$^3$School of Informatics, The University of Edinburgh \\
$^4$College of Computer Science and Technology, Zhejiang University \\ 
$^5$Alibaba-Zhejiang University Joint Research Institute of Frontier Technologies \\
$^6$Hangzhou Innovation Center, Zhejiang University}
\begin{document}

\maketitle

\begin{abstract}
Knowledge graph (KG) reasoning is becoming increasingly popular in both academia and industry. Conventional KG reasoning based on symbolic logic is deterministic, with reasoning results being explainable, while modern embedding-based reasoning can deal with uncertainty and predict plausible knowledge, often with high efficiency via vector computation. A promising direction is to integrate both logic-based and embedding-based methods, with the vision to have advantages of both. It has attracted wide research attention with more and more works published in recent years. In this paper, we comprehensively survey these works, focusing on how logics and embeddings are integrated. We first briefly introduce preliminaries, then systematically categorize and discuss works of logic and embedding-aware KG reasoning from different perspectives, and finally conclude and discuss the challenges and further directions.
\end{abstract}

\vspace{-3mm}
\section{Introduction}
Knowledge representation and reasoning plays a critical role in many domains, especially Artificial Intelligence(AI).
Knowledge graph (KG), representing facts in the form of triples, with vocabulary defined in a schema (also known as ontology),
is a simple yet efficient and increasingly popular way of knowledge representation. Many general-purpose and domain-specific KGs, such as Wikidata and SNOMED Clinical Terms are under fast development and widely used.
Reasoning, which aims to discover implicit knowledge, can significantly boost the usage of KGs in many applications such as question answering and assist KG curation by e.g., link prediction.
Conventional KG reasoning is often based on symbolic logics and is deductive. For example, HermiT~\cite{Hermit} is a classic description logic reasoner for OWL\footnote{Web Ontology Language. \url{https://www.w3.org/OWL/}} ontologies; RDFox~\cite{RDFox} is a famous KG storage supporting Datalog rule reasoning.
Recently, with the fast development of deep learning, 
KG embedding, which represents entities and relations as vectors (embeddings) with their relationships reflected, shows great success in KG reasoning, especially in inductive reasoning without pre-defined logics.
Logic-based reasoning is usually interpretable and transferable, while embedding-based reasoning can deal with uncertainty and data noise, and is able to predict non-determined but plausible knowledge. 
Thus flourishing research has been conducted to integrate logics and embeddings for more robust KG reasoning.

There are several survey papers about both KG and neural-symbolic integration.
\cite{SWJ-survey} focuses on combing symbolic reasoning with statistical reasoning, while \cite{AIOpen-survey} reviews papers on symbolic, neural, and hybrid reasoning.
Unlike these papers, our survey has a more specific topic, focusing on logics and KG embeddings,
and has a more fine-grained categorization to these studies, from two general perspectives: (i) injecting logics, such as logical rules and ontological schemas, into embedding learning, and (ii) utilizing KG embeddings for logic reasoning-relevant tasks, such as query answering, theorem proving and rule mining.
More importantly, we analyze and compare methods of each category from some more fine-grained perspectives, including logic types, the existence of pre-defined logics, integration stages, and integration mechanisms.
The survey also introduces the challenges and discusses the potential directions. All these are beneficial for future research on KG reasoning as well as KG construction, KG application, and neural-symbolic integration.

The remainder of the paper is organized as follows. Section 2 briefly introduces the background on logic-based and embedding-based KG reasoning. Section 3 reviews methods that utilize logics for augmenting embeddings. Section 4 reviews methods of using embeddings for supporting reasoning tasks. Section 5 analyzes the challenges and discusses the future directions. 
% The final section concludes the paper.}
%In section 2, we briefly introduce the background of reasoning based on  logics and embeddings, the basis of logic and embedding integration. Then in section 3, we comprehensively review methods of logics for emebddings, including logic rules for embeddings and ontological schemas for embeddings. Next we comprehensively review methods of embeddings for logics, including embedings for logic reasoning and embeddings for logic learning. After that, we briefly introduce the applications and benchmarks in Section 6. Finally, in Section 7, we outline challenges and future directions in this area. 

\vspace{-2mm}
\section{Background}
\vspace{-1mm}
\paragraph{Reasoning with Logics}
% reasoning language 
% probabilistic reasoning 
% rule learning
Given a KG $\mathcal{G}$, logical reasoning can be used to infer new implicit knowledge or to detect inconsistencies.
Many might be familiar with logical rules. 
One rule, which can be simply represented as  $\textit{H} \gets \textit{B}_1 \land  \textit{B}_2 \land ... \land \textit{B}_n$,  means that the head atom $\textit{H}$ can be inferred by the body atoms $\textit{B}_1$, ..., $\textit{B}_n$.
For example, \textit{isfatherOf(X, Y)}  $\gets$ \textit{Male(X)}  $\land$  \textit{isParentOf(X, Y)} means that if $X$ is a male, and $X$ is a parent of $Y$, then $X$ is the father of $Y$. In addition to rules, the  Web Ontology Language OWL 2, which is based on Description Logics (DLs), is a key standard schema language of KGs.  It is based on the $\mathcal{SROIQ}$ DL~\cite{HKS06}.  OWL 2 provides rich expressive power, including a strong support for datatypes~\cite{PaHo06} and rules~\cite{KRH08}. 
OWL 2 schema can be used to define class hierarchies, complex classes and relations, domain and range for relations, and more complex schema axioms. Logic languages allow deductive reasoning, such as consistency checking, materialization, query answering, as well as inductive and abductive reasoning.
%defining the hierarchy of entities and relations, as well as properties of relations
%such as \textit{Symmetric}, \textit{Reflexive} and \textit{Transitive} property. 
%Based on 
%\cjy{With} well defined \cjy{logics, KG tasks}
%logic forms, logic reasoning 
%such as structured query answering and theorem proving \cjy{can be augmented by implicit knowledge from deductive reasoning.
%Meanwhile, inductive reasoning, such as discovering logic formulae (a.k.a. logic learning), is also feasible and has been investigated. 
%could be conducted automatically, and could achieve high precision for formulas that are accurate. 
%To find accurate formulas, logic learning has also developed into an own research area to automatically mine logic formulae based on observed triples. 

\paragraph{Reasoning with Embeddings} 
%In the context of knowledge graph, embedding, referring to the representation of entities and relations in vectors space, is emerged with the development of Knowledge Graph Embedding (KGE) methods. 
KG embedding (KGE), as a kind of representation learning technique, aims to represent entities and relations by vectors with their semantics (e.g., relationships) preserved in the vector space. With the embeddings, implicit and new knowledge can be inferred, often with approximation or prediction.
Many successful KGE methods, such as TransE~\cite{TransE}, ComplEx~\cite{ComplEx} and RotatE~\cite{RotatE}, have been developed in the past decade.
A KGE method is usually composed of a score function $\phi()$ which defines how to compute the truth value based on entity and relation embeddings, and a loss function that maximizes the truth value of explicit positive triples and minimizes values of generated negative triples. 
Take TransE~\cite{TransE}  as an example, it makes $||\mathbf{h} + \mathbf{r} - \mathbf{t} ||$ as score function of triples, where $\mathbf{h}, \mathbf{r},$ and $\mathbf{t}$ are vectors of $h, r$, and $t$ in the Euclidean space, respectively. The embeddings are learned based on a margin loss that tried to maximize the margin between truth value of possitive and negative triples. 
Finally, KGEs can be used to query triples, infer new triples and discover inconsistency between triples.

% \section{Logics for Embedding}
% \section{Logic Guided Embedding Learning}
% \vspace{-1mm}
\section{Logics for Embeddings}
% \vspace{-1mm}
\label{sec:logics4embeddings}
% The core argument against embeddings is their supposed inability to capture deeper semantics and more complex patterns of reasoning such as first-order logic. 
% Rule-based reasoning could be easily extended to novel entities and relations and improved for existing but inaccurate ones. 
% Combining deep neural networks with structured logic rules is desirable to harness flexibility and reduce uninterpretability of the neural models. 
% Triples are represented in atomic formulae and rule are represented as complex formulae.

%Embedding-based methods embed relational facts in knowledge graphs into vector space where the calculation is highly efficient. 
Though KGE methods have achieved great success, they still suffer from the following problems: 1) they could fail to embed complex semantics such as relationships by logical rules; 2) they usually ignore the ontological schema of a KG as inputs; 3) most of them build black-box models with a shortage of explanation. Integrating logics into embeddings is expected to solve these problems. Before analyzing concrete methods of such KG embedding, we first introduce the perspectives that are considered in categorization and discussion: 
%While there are a few argument against embeddings including 1) their inability to capture deeper semantic and more complex patterns of reasoning such as logical rules; 2) they ignore encoding ontological schema since a KG is often composed of more than relational facts, e.g., many popular KGs such as DBpedia and NELL have ontological schemas;  3) they could not be easily extended to novel entities and relations; 3) they are one-step black-box model that are unexplainable. 
%Combining embedding-based methods with formal logics is desirable to harness expressiveness, robustness, and explainability of them. 
%
%
%There are mainly two types of logics injected for embeddings, logical rules and ontological schemas.
%Before reviewing methods for these two types of logics, we would like to first introduce several perspectives of analyzing methods of logics for emebddings:  
\paragraph{Logic Types} 
We consider two mainstream logic forms used in KG reasoning: (1) logic rules and (2) Ontological schema, such as those supported by OWL 2, which can internalize most logic rules, with some exceptions, such as those with loops in the body~\cite{KRH08}. 

\paragraph{Pre-defined logics} 
1) \textit{With}: injecting predefined logics or their reasoning results into embeddings, and/or applying ontological axioms as constraints. 
2) \textit{Without}: injecting the concept of logics into embeddings, where logics are assumed existing in the KG while no specific data of logics are used.

\paragraph{Integration Stages} 
Considering the time when logic is injected in learning embeddings, there are three stages:
1) \textit{Pre}: conducting symbolic reasoning before learning embeddings. The reasoning often impacts training samples such as positive and negative triples.
2) \textit{Joint}: injecting the logics during embedding learning. This often extends the loss function with logic constraints, by e.g., adding an additional regularizer on some relation embeddings.
3) \textit{Post}: conducting symbolic reasoning after embeddings are learned, by e.g., jointly constructing a predictive model with both results from embeddings and logics as inputs, or logical constraints as filters.
% \begin{itemize}
%     \item \textit{Pre}: conducting symbolic reasoning before learning embeddings. The reasoning often impacts training samples such as positive and negative triples.
%     \item \textit{Joint}: injecting the logics during embedding learning. This often extend the loss function  with logic constraints, by e.g., adding an additional regularizer for some relation properties.
%     \item \textit{Post-}: conducting symbolic reasoning after embeddings are learned, by e.g., jointly constructing a prediction model with both embeddings and   logics, or logical constraints as filters.
% \end{itemize}
\paragraph{Mechanisms} 
1) \textit{Data-based}: replacing variables in logic expressions with concrete entities and getting new triples, then adding all or part of the new triples into training. 
2) \textit{Model-based}: adding constraints on the embedding of entities and relations included in logic expressions into training. No additional triples are used. 
% Two typical mechanisms are:  
% \begin{itemize}
%     \item \textit{Data-based} Replacing variables in logic expressions with concrete entities and get new triples, then adding all or part of the new triples into training of embedding.   
%     \item \textit{Model-based} Adding constrains on embedding of entities and relations included in logic expressions into training of embeddings. No additional triples are used. 
% \end{itemize}

% According to types of logics, we introduce methods of logic rules and ontological schema for embeddings.  Based on the stage of injecting, there are three types, pre-learning injection, joint-learning injection, and post-learning injection. From technique point of view, there are grounding based and regularizing based methods. Based on the existence of logics, there are explicitly and implicitly injection. 
% Next we separately reviews KGE methods with logic rules and with ontological schemas. See Table~\ref{tab:summary1} for a summary.
%Next, we comprehensively review methods of logic rules for embeddings and ontological schemas for embeddings, 
% Next, we review methods mainly according to types of rules, 
%and discuss each method based on the perspectives of logic types, injecting ways, injecting statge and mechanism, as introduced before, and summarize them in Table~\ref{tab:summary1}.

% \subsection{Types of logic}
% % 

% \subsection{Pre- V.S. Joint- V.S. Post-training Injection}

% \subsection{Grounding-based V.S. Regularization-based}

% \subsection{Explicitly V.S. Implicitly}

\vspace{-1mm}
\subsection{Logic Rules for Embeddings}
\vspace{-1mm}
Logic rules are of the form $\textit{H} \gets \textit{B}_1 \land  \textit{B}_2 \land ... \land \textit{B}_n$ where \textit{H} is the rule head and $\textit{B}_1 \land  \textit{B}_2 \land ... \land \textit{B}_n$ is the rule body with the conjunction of atoms. A typical kind of logic rule is path rule where rule body is a path from a head variable to a tail variable in the rule head; 
for example, \textit{r(X,Y)}$\gets$  \textit{r}$_1$\textit{(X,Z)} $\land$ \textit{r}$_2$\textit{(Z,Y)} is a path rule with a path from variable $X$ to $Y$ as body.
Note that the number of atoms in the body is also known as the rule length.
There are a few works that inject path rules into embeddings. 
PTransE~\cite{PTransE} is a typical explicit and joint-training method, where path compositional representations calculated with relation embeddings in the path are encouraged to near the relation embedding in rule head in vector space. 
PTransE applies all high-quality paths during training. Alternatively, RPJE~\cite{RPJE} selects the path with the highest confidence to compose the path for each triple and distinguishes paths with length $1$ from other paths, for which confidence of rules are not considered during training. 
Instead of compositional representation regularization, ComplEx-NNE\_AER~\cite{ComplEx-NNE_AER} infers constrains on relation embeddings through making $\phi(h,r_1, t) > \phi(h, r_2, t)$ if $\textit{r}_1\textit{(X,Y)} \gets \textit{r}_2\textit{(X,Y)}$, and  the larger $\phi(h,r, t)$ is, the more possible $(h,r,t)$ to be a positive one. Following ComplEx-NNE\_AER, SLRE~\cite{SLRE} adapts relation embedding constrains to longer path rules. 

Methods mentioned above inject pre-defined rules via regularizing relation embeddings during training. 
They are specific to KGE methods.
One more general solution is using grounding, which replaces variables in each rule with concrete entities, infers implicit triples, and generates additional triples for KGEs' training.
For example, KALE~\cite{KALE} models groundings by t-norm fuzzy logics which  gives a truth score for each grounding based on truth value of all atoms. It trains groundings with negative sampling together with existing triples.  
KALE conducts grounding before training and injects them one time. Alternatively, multiple times or iterative injection ensures more flexibility for a model. For example, in each training iteration, RUGE~\cite{RUGE} and IterE~\cite{IterE} predicts labels of unlabeled triples in groundings based on t-norm fuzzy logics, and uses labeled triples and post-labeled triples for training. The iterative manner enables the model to predict labels for unlabeled triples dynamically based on embeddings. 
These data-based methods need to materialize each rule, resulting in a massive number of groundings, thus they do not scale well to large KGs. However, the KGE-free property makes them continuously benefit from the development of KGEs. 
Post-training injection solutions are also KGE-free. 
For example, \cite{DBLP:conf/ijcai/WangWG15} proposes to frame an Integer Linear Programming problem to combine rules and KGEs, where rules are translated into conditional constraints during training, and scores from well-trained KGEs are inputs. 

Apart from the satisfaction of rules, TARE~\cite{TARE} emphasizes the properties of transitivity and asymmetry of rules which makes the order of relations in rules matter, and it models the order of relation types in logic rules by the component-wise inequality.

\begin{table*}[]
    \centering
    \resizebox{\textwidth}{!}{
    \begin{tabular}{c | c c | c c c| c c | c}
    \toprule 
       \multirow{2}{*}{\textbf{Logic Types}} & \multicolumn{2}{c|}{\textbf{Pre-defined Logics}} & \multicolumn{3}{c|}{\textbf{Integration Stage}} & \multicolumn{2}{c|}{\textbf{Mechanism}} & \multirow{2}{*}{\textbf{Methods}}  \\
        & \textit{With} & \textit{Without} & \textit{Pre} & \textit{Joint} & \textit{Post} & \textit{Data-based}  & \textit{Model-based} &  \\
        \midrule 
        \multirow{3}{*}{path rule} & \multirow{3}{*}{$\surd$} & \multirow{3}{*}{-} & \multirow{3}{*}{-} & \multirow{3}{*}{$\surd$} & \multirow{3}{*}{-} & \multirow{3}{*}{-} & \multirow{3}{*}{$\surd$} & \cite{PTransE,SePLi,RPJE} \\
        & & & & & & & &  \cite{RPJE,ComplEx-NNE_AER}    \\
        & & & & & & & &  \cite{SLRE,TARE}\\
        path rule & $\surd$ & - & $\surd$ & - & - & $\surd$ & - & \cite{KALE,RUGE} \\
        path, cardinality & $\surd$ & - & - & - & $\surd$ & - & $\surd$ & \cite{DBLP:conf/ijcai/WangWG15} \\
        class hierarchy & $\surd$ & - & - & $\surd$ & - & - & $\surd$ & \cite{TKRL} \\
        class hierarchy & - & $\surd$ & - & $\surd$ & - & - & $\surd$ & \cite{HAKE} \\
        entity type & $\surd$ & - & - & $\surd$ & - & - & $\surd$ & \cite{TAGAT,TypeKGE}\\
        entity type & $\surd$ & - & - & $\surd$ & - & $\surd$ & $\surd$ & \cite{RETA}\\
        entity type & - & $\surd$ & $\surd$ & $\surd$ & - & $\surd$ & $\surd$ & \cite{IterefinE} \\
        % IterE? 
        relation hierarchy & - & $\surd$ & - & $\surd$ & - & - & $\surd$ & \cite{HRS,HAKE} \\
        domain, range & $\surd$ & - & $\surd$ & - & - & $\surd$ & - & \cite{TRESCAL} \\
        domain, range & - & $\surd$ & - & $\surd$ & - & - & $\surd$ & \cite{TypeKGE} \\
        equivalent, inverse & $\surd$ & - & - & $\surd$ & - & - & $\surd$ & \cite{KGE_R}\\
        ansymmetric & - & $\surd$ & - & $\surd$ & - & - & $\surd$ & \cite{ComplEx}\\
        composition & - & $\surd$ & - & $\surd$ & - & - & $\surd$ & \cite{RotatE}\\
        transitive & - & $\surd$ & - & $\surd$ & - & - & $\surd$ & \cite{Rot-Pro}\\
        reflexive, symmetric, transitive & - & $\surd$ & - & $\surd$ & - & - & $\surd$ & \cite{dORC} \\
          ontological schema & $\surd$  & - & $\surd$ & - & $\surd$ & $\surd$ & - & \cite{SIC} \\
        \bottomrule
    \end{tabular}
    }
    \vspace{-2mm}
    \caption{Summary of methods injecting logics into KG embeddings.}
    % \vspace{-3mm}
    \label{tab:summary1}
\end{table*}
% \vspace{-1mm}
\subsection{Ontological Schemas for Embeddings}
% \vspace{-1mm}
% \subsection{Web Ontology Language}
%Ontological Schema defines the higher level knowledge about entities and relations, such as hierarchy and properties of them. 
Ontological schemas, which are often defined by languages such as OWL and RDF Schema, describe high-level semantics (meta information) of KGs.
We next survey methods that inject class hierarchies, relation hierarchies, and relation properties.
%such as entity classes (types), class hierarchies, relations, relation hierarchies, and relation properties. 
% Due to diversity of ontological schemas, 
%Next we comprehensively survey methods injecting three types of schemas, class hierarchies, relation hierarchies and relation properties, classified from semantic point of view. 

\paragraph{Class Hierarchies} 
Class hierarchies classify entity types, denoting entities as instantiations of classes. There are two tasks for injecting class hierarchies, encoding the types of entities and encoding hierarchies of entity types. 

To encode entity types, with entity types given, one kind of method learns an embedding for each type, and adds regularization one these embeddings. For example, 
TAGAT~\cite{TAGAT} 
regularizes entity embeddings to be close to their corresponding type embeddings, and also closes the embeddings of entities that belong to the same type. 
RETA-Grader~\cite{RETA} uses type embeddings in entity-typed triples  $(h\_type_i, r, t\_type_j)$ for each triple $(h,r,t)$, and  concatenate embeddings in these two types of triples as inputs for triple scoring.  
Without entity types given, a common way is to assume a type for each entity and learn an embedding for it. For example,  TypeDM~\cite{TypeKGE} uses the assumed type embedding of an entity and domain(range) embeddings of a relation to calculate the satisfactory of domain(range) of a relation. 
Another way is to infer candidate types of each entity based on ontological information. For example, IterefinE~\cite{IterefinE} applies the inferred types to refine the KG data and regularize the learning of embeddings.  
% IterefinE~\cite{IterefinE} uses ontological information and inference rules to infer candidate types of each entity, and applied TypeE, a method similar to ~\cite{TypeKGE}, to learn from typed entities and triples, where not only type information from PSL-KGI is utilized in TypeE, but also refined KG through the predicted triples from TypeE is feedback to PSL-KGI. 

% To inject entity hierarchy, 
% TKRL~\cite{TKRL} proposes three hierarchical type encoders to generate type projection matrix for entities including general from of type encoder, averaging class matrices, recursive hierarchy encoder and weighted hierarchy encoder. For each entity, there are multiple representations in different types.
In order to inject entity type hierarchies into embeddings, there are methods with and without pre-defined hierarchies.
With class hierarchies given, one kind of method combines the type hierarchy of each entity to its embedding. For example, TKRL~\cite{TKRL} encodes type hierarchies into projection matrix, and injects type hierarchies into entity embeddings via projecting them with projection matrices. 
Without pre-defined class hierarchies, HAKE~\cite{HAKE} proposes to map entities into polar coordinate system, where concentric circles can naturally reflect hierarchies.
After training, implicit entity hierarchies could be decoded from entity embeddings.

% Above mentioned methods explicitly inject ontological axioms that exist. There are also methods conduct implicitly injection that design embedding models following the instruction of semantics of ontological schema. 
 
% TypeDM/Complex~\cite{TypeKGE} proposed to make type information into consideration of knowledge graph embedding via a type-sensitive knowledge base inference without explicit type supervision. It learns an additional two embeddings for each relation to indicate the domain and range of it and learns one additional embedding for each entity for its type. The domain and range satisfactory are calculated via dot product between corresponding embeddings. 

% proposes to incorporate embeddings with entity type information via generating entity-typed triples like $(h\_type_i, r, t\_type_j)$  and learning a schema relatedness feature vector using similar methods as learning from normal triples. And finally triple and schema relatedness feature vectors will be concatenated into an overall relatedness feature vector for prediction.
% IterefinE~\cite{IterefinE} interatively iteratively combines PSL-KGI which uses ontological information and inference rules to infer candidate types of each entity, and TypeE method similar to ~\cite{TypeKGE} to learn from typed entities and triples, where not only type information from PSL-KGI is utilized in TypeE, but also refined KG through the predicted triples from TypeE is feedback to PSL-KGI. 

\paragraph{Relation Hierarchies}
Relation hierarchies contain subsumption relationships between relations; 
for example, \textit{hasFather} is a sub-relation of \textit{hasParents}. 
Without pre-defined relation hierarchies, 
HRS~\cite{HRS} assumes a three-layer hierarchical relation structure for each relation, including relation clusters, relations and sub-relations, discovered by the K-means algorithm on relation embeddings from KGE.  
To encode relation hierarchies, it learns an embedding for each relation cluster and represents relations as the sum of their cluster embedding, relation embedding, and sub-relations' embedding. 
Another method TransRHS~\cite{TransRHS} which does not explicitly assume a multi-layer relation hierarchy, proposes to model each relation as a vector together with a relation-specific sphere. 
It assumes lower-level relations are with smaller spheres. If a predicted entity lies in the spheres of a lower level relation such as \textit{hasFather}, then the model will ensure it also lies in the sphere of its parent relations such as \textit{hasParent}. This embodies the inherent generalization relationships among relations. 

\paragraph{Relation Properties}
%A lot ontological schemas defines the properties of relations are explored to be injected into embedding. 
% \textit{Domain} and \textit{Range} defines the suitable type of head and tail entity for one relation, respectively. 
% To inject properties of relations, 

%For a relation, there are multiple ontological schemas include only this relation, 
Ontological schemas often define quite a few relation properties (and constraints). 
% Some of them have been considered and injected into learning KG embeddings.

First, we introduce properties constraining only one relation that have been considered. 
For \textit{domain} and \textit{range} of relations, TRESCAL~\cite{TRESCAL} leverages them by filtering triples in KGs where entities are not compatible with the domain or range of relations.  
If not pre-defined, TypeDM~\cite{TypeKGE} learns an assumed domain and range embedding for each relation and uses them for constraining entity type embeddings. 
To model \textit{Ansymmetric} relations, ComplEx~\cite{ComplEx} proposes to
embed KGs in complex vector space, where the commutative property for multiplication is not satisfied. 
To further model \textit{Composition}\footnote{Composition of relations includes multiple relations, we introduce it in this paragraph because RotatE is based on ComplEx.} between relations,  RotatE~\cite{RotatE} proposes to define each relation as a rotation from the head entity to the tail entity in complex vector space. 
Furthermore, Rot-Pro~\cite{Rot-Pro} proposes to model \textit{Transitive} relations by defining relations as projection and rotation. 
dORC~\cite{dORC} enables modeling \textit{Reflexive}, \textit{Symmetric} and \textit{Transitive} relations by disentangling the embedding of each entity as head and tail entities.
\cite{DBLP:conf/ijcai/WangWG15} propose a post-training injection for \textit{Cardinality} of relations, by framing a Liner programming problem.
% ~\cite{DBLP:conf/ijcai/WangWG15} proposes to incorporate them, speficically referring to at-most-one-restraint on number of tail entities of on relations, 
% by framing a Liner programming problem, to achieve post-training injection. 

Second, we introduce relation properties constraining multiple relations that have been considered, including \textit{Equivalent} and \textit{Inverse}. 
Given these two pre-defined relation properties, \cite{KGE_R} injects them via a single constraint on the embedding of the two relations in schemas, based on the vector space assumption in KGEs.
Besides applying axiom-based regularization, TransOWL~\cite{TransOWL} also proposes to add new triples inferred by these schemas into training, which could also be applied to entity type constraints such as \textit{equivalentClass} and \textit{subClassOf}.  
More generally, SIC~\cite{SIC} proposes to use ontological reasoning within their iterative KG completion approach to inject inferred triples to enrich the input KG for embedding, and to filter out schema-incorrect triples via consistency and constraint checking. ReasonKGE~\cite{JTGS2021} follows SIC~\cite{SIC} by using schema-incorrect triples for negative sampling.   

\vspace{-1mm}
\subsection{Summary}
\vspace{-1mm}
We summarize methods injecting logics into embedding methods introduced before in Table~\ref{tab:summary1}. 
If logics are pre-defined, there are 
diverse methods with different integration stages and mechanisms. Moreover, in the scenario without pre-defined logics, model-based and post-training injection methods are used. For pre-training injection methods, data-based models are more common, while model-based models are more common for joint-training injection. We separately discussed injecting logical rules and ontology schemas for embeddings in this section, considering they are two types of logic languages. However, they are not entirely distinct, and each ontological schema could be rephrased into a logic rule. For example composition of relations could be represented as a path rule. 

\vspace{-1mm}
\section{Embeddings for Logics}
\vspace{-1mm}
\label{sec:embeddings4logics}
% \section{Logic Reasoning and Learning with Embeddings}
%Formal logic reasoning methods are incredibly powerful at achieving high precision for formulas that are accurate, and the logical representations of facts and rules can be fed to automated theorem prover to perform new fact inference as well as inconsistency checking. 
Pure symbolic reasoning with different kinds of logics have been investigated for years and widely applied, but it still suffers from the following problems:
1) the logics must be given as a priori, but constructing logics often relies on domain experts, costing a lot of time and labor, and the logics in most applications are underspecified, limiting the knowledge that can be inferred;
2) it often cannot cope with noise and uncertainty inherent to real-world data; 
3) logic reasoning often cannot scale up since some (complex) logics may lead to high time or space complexity.
In contrast, embedding-based reasoning is good at inductive reasoning without pre-defined logics, can well address uncertainty, and can scale up by approximation.
Thus utilizing embeddings for augmenting KG reasoning tasks attracts wide research attention.
We mainly review two kinds of KG reasoning tasks: deductive logic reasoning, which further includes query answering and theorem proving, and inductive logic learning.
Before discussing the methods, as in Section~\ref{sec:logics4embeddings}, we first introduce the perspectives for analyzing and categorization:

\paragraph{Logic Types} Many logics are considered in each work, such as path rules, numerical rules, path queries, and logic queries constructed by $\land$, $\lor$, $\lnot$, and $\ne$. We present logic types following the expression in logical forms they originated. 

\paragraph{Pre-defined Logics} 
Embeddings could be combined to logics 1)\textit{With} and 2)\textit{Without} pre-defined logics. 

\paragraph{Combination Stage} Considering the time when embeddings are combined to logics, there are two stages:
1) \textit{Pre}: applying embeddings before logic reasoning or learning, such as for candidate selection. 
2) \textit{Joint}: applying embeddings during logic reasoning or learning.

\paragraph{Mechanism} 
1) \textit{Hybrid}: after applying embeddings, methods still infer in a symbolic space. 
2) \textit{Neural}: using embeddings following the process of logic reasoning, and all inferences are conducted in vector space.

\begin{table*}[]
    \centering
    \resizebox{\textwidth}{!}{
    \begin{tabular}{c|c | c c | c c | c c | c}
    \toprule 
        \multirow{2}{*}{\textbf{Tasks}} & \multirow{2}{*}{\textbf{Logic Types}} & \multicolumn{2}{c|}{\textbf{Pre-defined Logics}} & \multicolumn{2}{c|}{\textbf{Integration Stage}} & \multicolumn{2}{c|}{\textbf{Mechanism}} & \multirow{2}{*}{\textbf{Methods}}  \\
        & & \textit{With} & \textit{Without} & \textit{Pre} & \textit{Joint} & \textit{Hybrid} & \textit{Neural} & \\
        \midrule 
        Query Answering &  path query & $\surd$ & - & - & $\surd$ & - & $\surd$ & \cite{traversing} \\
        Query Answering & path query & $\surd$ & - & $\surd$ & - & $\surd$ & - & \cite{INS-ES} \\
        Query Answering & $\land$ & $\surd$ & - & - & $\surd$ & - & $\surd$ & \cite{GQE,BiQE} \\
        Query Answering & $\land$, $\lor$ & $\surd$ & - & - & $\surd$ & - & $\surd$ & \cite{Query2box,CQD} \\
        Query Answering & $\land$, $\lor$, $\lnot$ & $\surd$ & - & - & $\surd$ & - & $\surd$ &\cite{BetaE,ConE} \\
        Query Answering & $\land$, $\lor$, $\lnot$, $\ne$ & $\surd$ & - & - & $\surd$ & - & $\surd$ & \cite{NewLook}\\
        Theorem proving & path rule & $\surd$ & $\surd$ & - & $\surd$ & $\surd$ & - & \cite{NTP} \\
        Theorem proving & path rule & $\surd$ & $\surd$ & - & $\surd$ & $\surd$ & - & \cite{GNTP,CTP} \\
        Rule Mining & path rule & - & $\surd$ & $\surd$ & $\surd$ & $\surd$ & - & \cite{RuLES,RLvLR,ProPPR+MF} \\
        Rule Mining & path rule & - &$\surd$ & - & $\surd$ & - & $\surd$ & \cite{NeuralLP,DRUM}\\
        Rule Mining & numerical rule & - &$\surd$ & - & $\surd$ & - & $\surd$ & \cite{Neural-Num-LP}\\
        \bottomrule
    \end{tabular}
    }
    \vspace{-2mm}
    \caption{Summary of methods integrating embeddings for symbolic logic reasoning and learning.}
    % \vspace{-3mm}
    \label{tab:summary2}
\end{table*}

% \subsection{Symbolic Reasoning}
\vspace{-1mm}
\subsection{Embeddings for Logic Reasoning}
\vspace{-1mm}
Query answering and theorem proving are two popular logic reasoning tasks where embeddings could be utilized. They are originally implemented by pure deductive symbolic reasoning with pre-defined logics. With the embeddings, additional knowledge can be predicted for more robust results.

\paragraph{Query Answering}
Query answering returns correct entities in a KG as answers of a given structured query, where reasoning is usually considered for hidden answers. Conventional query answering is conducted based on structure query languages such as SPARQL\footnote{\url{https://www.w3.org/TR/rdf-sparql-query/}} to retrieve and manipulate knowledge in a KG.

%Logic reasoning methods are also vulnerable to the incompleteness of and noise in KGs. Thus query answering with help of embeddings is widely researched to keep the ability of handling complex queries on incomplete or noise KGs. 
%where logic queries are pre-defined and the answers are given according to calculation in vector space. 
Quite a few studies use embeddings to KG incomplete and noise in query answering.
In the beginning, simple queries are considered, for example, path queries proposed in \cite{traversing}. 
It interprets TransE as implementing a soft edge traversal operator and recursively applies it to predict compositional path queries and is trained on path samples from random walks and explicit triples. 
Apart from simple path queries, 
more complex queries, such as conjunctive logical queries and Existential Positive First-Order (EPFO), involving multiple unobserved edges, nodes, and even variables are also widely researched with the help of embeddings. 
GQE embeds entities as a vector, relations as projection operators on entity embeddings, and makes $\land$ in conjunctive logical queries as intersection operators. 
Through these embeddings and operators, 
it encodes each query into a vector and gives answers based on the similarity between query and candidate entity embeddings.
BiQE~\cite{BiQE} translates conjunctive queries into a sequence and encodes them by Transformer Encoder. 
Query2box~\cite{Query2box} can further support disjunctions ($\lor$) in queries via transforming them into Disjunctive Normal Form (DNF), and it defines vector space operators for each type of quantifications. 
Alternatively, CQD~\cite{CQD} only defines projection operators using ComplEx~\cite{ComplEx} while applying other quantifications according to t-norms. 
To support a complete set of first-order logical operations, including conjunction($\land$), disjunction($\lor$) and negation($\lnot$), BetaE~\cite{BetaE} and ConE~\cite{ConE} propose to embed entities and queries as Beta distributions and sector-cones respectively, on which projection, intersection and negation  operator are defined. 
NewLook~\cite{NewLook} further supports queries including Difference($\ne$) by logical operations as flexible neural networks. 
Some studies use embeddings to improve the efficiency of query answering, especially for those queries with complex logics.
For example, INS-ES~\cite{INS-ES}, running the data-driven inference algorithm INS on Markov Logic Network (MLN) for symbolic reasoning, uses embeddings from TransE to generate a much smaller candidate set for subsequent fact inference in INS. 
% Above mentioned query answering methods adopt \todo{joint-reasoning combination of embeddings} where queries are answered in vector space with embeddings.
% INS-ES~\cite{INS-ES} takes a\cjy{n} alternative way of pre-reasoning combination. It runs the data-driven inference algorithm on Markov Logic Network (MLN) called Inferring via Grounding Network Sampling (INS) for symbolic reasoning, in which embeddings are used for instance selection, generating a much smaller candidate sets for subsequent fact inference in INS. It first employs TransE~\cite{TransE} to learn the representations of entities and relations in the KG, then calculates similarity scores between candidates and the input query, and selects the top-N instances to constitute a new smaller candidate set for subsequent fact inference. 

\paragraph{Theorem Proving}
Another task of logic reasoning is theorem proving, automatically inferring triples given a set of facts and predefined logic rules. 
Conventional theorem proving methods are based on different logic languages, such as Prolog, Datalog, and OWL, which are vulnerable to incomplete and noise KGs.
Differentiable theorem proving using embeddings overcome the limits of symbolic provers on generalizing to queries with similar but not identical symbols.
With NTP~~\cite{NTP} as an example, 
it enables Prolog to learn embeddings and similarities between entities and relations in a KG. 
It keeps the variable binding symbolic following the inference process of Prolog but compares symbols using their embeddings rather than identical symbols. 
It could learn without predefined domain-specific rules and seamlessly reason with them. 
The core process of NTP follows steps of symbolic logic reasoning that requires enumerating and scoring all bounded-depth proof paths for a given goal, thus NTP  is inefficient on large KGs.   
Thus in some works, embeddings are also used to improve the efficiency of the differentiable prover. 
For example, GNTP~\cite{GNTP} use fact embeddings to select the top nearest neighbor facts for proving sub-goals, and also use relation embeddings to select top rules to be expanded. 
Another method CTP~\cite{CTP} uses a key-value memory network, conditioned on the goal to prove and embeddings of relations and constants, to dynamically generate a minimal set of rules to consider at each reasoning step.  

% CTP~\cite{CTP} is another extension of NTP that is able to learn an adaptive strategy for selecting subsets of rules to consider at each step of the reasoning process. CTP dynamically generate a minimal set of rules via a neural network architecture $\texttt{Select}$ module conditioned on the goal to prove and embeddings of relations and constants. Existing rule are regarded as useful prior for $\texttt{select}$ module and key-value memory network is used for rule selection and rules are stored in a differentiable memory. 

% Multi-hop 
% Query Answering
% differentiable Prover
% Markov Logic Network

% \vspace{-1mm}
\subsection{Embeddings for Logic Learning} 
% \vspace{-1mm}
% Rule learning 
% ontology learning
% Inductive logic programming~\cite{ILP}
Logic learning is to learn patterns from KGs and discover potential (and probabilistic) logics such as schemas and logic rules.
Conventional methods like AMIE~\cite{AMIE} and AnyBURL~\cite{AnyBURL} are symbolic-based. They 
determine structures of rules via random walking and adding atoms based on KGs, and measure the quality of rules by statistical matrices such as Confidence and Head Coverage. 
While statistical matrices might be misleading due to the incompleteness of and noise in KGs, thus it is difficult to learn high-quality rules from the explicit triples alone.
Embeddings are widely used in logic learning to overcome incompleteness and noise issues.
RuLES~\cite{RuLES} adds confidential triples using embedding models for quality extension of KGs. 
It iteratively extends rules induced from a KG through feedback from embedding models and evaluates the quality of rules on the origin KG and extended KG.
ProPPR+MF~\cite{ProPPR+MF} reconstructs the transition of proofs in ProPPR~\cite{ProPPR}, a stochastic extension of Prolog, based on embeddings of first-order logics through matrix factorization. 
% ProPPR computes a set of candidate formulas from KGs based on a structural gradient method.
% Based on the candidate formulas computed from ProPPR, ProPPR+MF constructs a graph between intermediate states of the proof and first-order logic formula and apples matrix factorization on the graph to learn first-order logics' embeddings. 
% Finally it transforms embeddings to parameters for the transition in the formula, which implicitly completes KGs.

%Next, we introduce methods using embeddings to improve the efficiency of rule learning. 
Embeddings have also been utilized to improve the efficiency of rule learning.
RLvLR~\cite{RLvLR} uses embeddings to guide and prune the search during rule mining, where embeddings are learned on the subgraph sampled for target predicates with RESCAL~\cite{RESCAL}. 
While DistMult~\cite{DistMult} and IterE~\cite{IterE} alternatively use relation embeddings to calculate the confidence of rules.
Another type of method is differentiable rule mining, learning rules in an end-to-end differentiable manner in vector space. 
They are inspired by TensorLog~\cite{Tensorlog}, a differentiable probabilistic logic. Tensorlog establishes a connection between inference using first-order rules and sparse matrix multiplication, and enables certain types of logical inference tasks to be compiled into a sequence of differentiable numerical operations on matrices.
Differentiable rule mining methods use a module containing relation embeddings to learn the weights for each operation of a rule used in TensorLog, for example, NeuralLP~\cite{NeuralLP} uses an attention-based neural controller system for weight generation, and DRUM~\cite{DRUM} applies a low-rank approximation.   
Besides path rules, Neural-Num-LP~\cite{Neural-Num-LP} further enable mining rules with numerical features in a differentiable framework. 

\vspace{-1mm}
\subsection{Summary}
\vspace{-1mm}
We summarize methods applying embeddings for logic reasoning and learning introduced before in Table~\ref{tab:summary2}. 
Embeddings could be combined with pre-defined logics either before(\textit{Pre}) or during(\textit{Joint}) the process of logic reasoning and learning. And  without pre-defined logics, embeddings usually are applied jointly. 
If the integration stage of embeddings is \textit{Joint}, there are both hybrid and neural methods, and only hybrid methods for \textit{Pre}. Methods that combining embeddings in a \textit{Post} manner haven't been proposed yet. 

\section{Conclusion and Discussion}
This paper presents a literature review for KG reasoning with logics and embeddings. We divided the integrated methods into integrating logics(logic rules and ontological schemas) to embeddings and integrating embeddings to logic-based reasoning tasks(query answering, theorem proving, and rule mining). Moreover, we analyzed those methods from four perspectives: logic types, the existence of pre-defined logics, integration stages, and integration mechanisms.

Next, we discuss challenges on this topic from four perspectives:  logic diversity, explainability, benchmark, and application. 
%logic and embedding integration, shared for logics for embeddings and embeddings for logics, from four perspectives. They are support of diverse logics , explainability, benchmark and application.
% In this section, we first discuss challenges and further directions of logics for embeddings and embeddings for logics, and then discuss from a broader view including benchmarks and applications. 

\paragraph{Logic diversity}
One critical challenge of logic and embedding integration lies in the diversity of logics.
The majority of current methods only consider specific kinds of logics, such as path rules, entity types (classes), class hierarchies, relation hierarchies, and relation properties (see Section~\ref{sec:logics4embeddings}), or specific logic quantifications such as conjunction, disjunction, and negation (see Section~\ref{sec:embeddings4logics}).
Only a few methods, like SIC, ReasonKGE, and TransOWL, support general ontological schemas.
Since a KG is often equipped with different kinds of logics, e.g., rules or ontological schemas, it becomes a significant challenge to support and integrate all or most of them simultaneously.

To this end, we think there are two promising directions. One is to explore more logic forms, such as rules with constants, universal quantification ($\forall$), disjointness, and more. 
Another is to research general frameworks that can simultaneously support different kinds of logics and is independent of KG embedding methods.
% Such frameworks could be implemented as disentangled embeddings with different logics injected in different components.
%, and different components could be used for different applications. 
% And framework could be independent to specific embedding methods, and continuously benefit from new methods proposed in the independent research area on KGEs. 

\paragraph{Explainability}
Another critical challenge is making integrated methods more explainable.
Most methods in  Section~\ref{sec:logics4embeddings} focus on improving the model's expressiveness, which does not change the black-box nature of embedding methods. 
Methods in Section~\ref{sec:embeddings4logics} enabling logic reasoning in vector space decrease the transparency of logic-based methods because the intermediate results are represented as embeddings that is understandable by human if and only if the meaning of these embeddings are properly interpreted, which is difficult in most cases. While for further AI applications, systems with higher safety, trustness, and fairness are expected. Improving the transparency of integrated methods could ensure them broader applications in the further.

In order to achieve this, potential directions may lie in the following three directions. 
First, for black-box models such as KGEs, inject not only semantics of logics but also reasoning steps into models, which extends them from one-step to multi-step reasoning models with interpretable intermediate results. 
Second, for multi-step models with embeddings as intermediate results, improve the interpretability of intermediate embeddings, especially generate a set of symbolic representations that is easy for humans to understand, instead of similarity intuitions between embeddings. 
Third, integrate logics and embeddings through explainable machine learning methods to help preserve the transparency of logic-based methods and the robustness of embedding-based methods.

\paragraph{Benchmark}  There is a shortage of resources for evaluating KG reasoning with both logics and embeddings.
Commonly used benchmarks for KGEs' evaluation, such as  WN18RR, FB15k-237 and NELL, are subsets sampled from one or multiple domain in large KGs.
When creating these datasets, the primary goal is making them suitable for supervised learning setting, which do not ensure and explore diverse logic patterns contained in the dataset. While many works injecting logics in embeddings or combining embeddings for logic reasoning are evaluated on these datasets, which could not reflect the capability of methods once the logic patterns they concerned are missing in the dataset. 
% \todo{In works of integrating embeddings in logic reasoning, new datasets sampled from existing ones that contain the specific logic patterns the work is concerning are usually created, such as FB15K-237-CQ, WN18-RR-PATH, which is not general for other works' evaluation.}
Therefore, benchmarks containing diverse pre-defined logics or logic patterns in triples deserved to be proposed.

Also, SIC~\cite{SIC} argues that experimental results show that the existing correctness notion based on the silver standard is highly questionable. Good results from the silver standard often cannot be transferred to other knowledge graphs beyond the benchmark KGs reported in papers. Instead, in the absence of large scale human evaluations,  schema correctness~\cite{SIC} is more promising. 

\paragraph{Application} 
% KGs are always involving, and thus reasoning with involving KGs, i.e., out-of-KG reasoning tasks, are quite important. 
%out-of-KG tasks are underexploitation for logic and embedding integrated methods. 
%Besides in-KG tasks like link prediction, query answering, triple classification, and rule mining, many out-of-KG tasks have been explored combining either logics defined in KGs or embeddings learned from KGs, such as relation extraction~\cite{LFDS}, recommender system~\cite{KGPretrain4RS}.
%, and image classification~\cite{OntoZSL}.
Apart from KG reasoning tasks, there are many KG applications, which have been proved could benefit from logics(embeddings) in KGs, for example, information extraction~\cite{LFDS}, recommender system~\cite{KGPretrain4RS}, and image classification~\cite{OntoZSL}, especially low-resource learning~\cite{chen2021low}. 
However, these tasks have been rarely explored with both logics and embeddings. 
Recently, inspired by the concept of pre-training and fine-tuning of language models, which is powerful in many downstream natural language processing tasks, KGEs are also extended to pre-trained KG models~\cite{PKGM} to be used in many KG applications. Thus in the future, applying both logics and embeddings into more KG applications and pre-trained KG models deserves attention. 
% logic and embedding integrated methods are expected to applied on  more independent out-of-KG task supported by KGs and pre-trained KG models. 
% challenges 和 discussion 穿插写
% including benchmark discussion
% 1 page

% further directions: 
% Integrating logic reasoning and machine learning could lead to more interpretable AI systems, leading higher safety, trustness and fairness, all of which are highly required in many applications.
% explainable 
% safety
% trustness
% 

\bibliographystyle{named}
\bibliography{ijcai22}

\begin{thebibliography}{}

\bibitem[\protect\citeauthoryear{Arakelyan \bgroup \em et al.\egroup
  }{2021}]{CQD}
Erik Arakelyan, Daniel Daza, Pasquale Minervini, and Michael Cochez.
\newblock Complex query answering with neural link predictors.
\newblock In {\em {ICLR}}, 2021.

\bibitem[\protect\citeauthoryear{Arora \bgroup \em et al.\egroup
  }{2020}]{IterefinE}
S.~Arora, S.~Bedathur, M.~Ramanath, and D.~Sharma.
\newblock Iterefine: Iterative {KG} refinement embeddings using symbolic
  knowledge.
\newblock In {\em {AKBC}}, 2020.

\bibitem[\protect\citeauthoryear{Bordes \bgroup \em et al.\egroup
  }{2013}]{TransE}
Antoine Bordes, Nicolas Usunier, Alberto Garcia-Duran, Jason Weston, and Oksana
  Yakhnenko.
\newblock Translating embeddings for modeling multi-relational data.
\newblock In {\em NIPS}, 2013.

\bibitem[\protect\citeauthoryear{Chang \bgroup \em et al.\egroup
  }{2014}]{TRESCAL}
K.~Chang, W.~Yih, B.~Yang, and C.~Meek.
\newblock Typed tensor decomposition of knowledge bases for relation
  extraction.
\newblock In {\em {EMNLP}}, 2014.

\bibitem[\protect\citeauthoryear{Chen \bgroup \em et al.\egroup
  }{2021}]{chen2021low}
Jiaoyan Chen, Yuxia Geng, Zhuo Chen, Jeff~Z Pan, Yuan He, Wen Zhang, and so~on.
\newblock Low-resource learning with knowledge graphs: A comprehensive survey.
\newblock {\em arXiv preprint arXiv:2112.10006}, 2021.

\bibitem[\protect\citeauthoryear{Cohen \bgroup \em et al.\egroup
  }{2020}]{Tensorlog}
W.~W. Cohen, F.~Yang, and K.~Mazaitis.
\newblock Tensorlog: {A} probabilistic database implemented using deep-learning
  infrastructure.
\newblock {\em J. Artif. Intell. Res.}, 2020.

\bibitem[\protect\citeauthoryear{d'Amato \bgroup \em et al.\egroup
  }{2021}]{TransOWL}
Claudia d'Amato, Nicola~Flavio Quatraro, and Nicola Fanizzi.
\newblock Injecting background knowledge into embedding models for predictive
  tasks on knowledge graphs.
\newblock In {\em {ESWC}}, 2021.

\bibitem[\protect\citeauthoryear{Ding \bgroup \em et al.\egroup
  }{2018}]{ComplEx-NNE_AER}
Boyang Ding, Quan Wang, Bin Wang, and Li~Guo.
\newblock Improving knowledge graph embedding using simple constraints.
\newblock In {\em {ACL}}, 2018.

\bibitem[\protect\citeauthoryear{Gal{\'{a}}rraga \bgroup \em et al.\egroup
  }{2015}]{AMIE}
L.~Gal{\'{a}}rraga, C.~Teflioudi, K.~Hose, and F.~M. Suchanek.
\newblock Fast rule mining in ontological knowledge bases with {AMIE+}.
\newblock {\em {VLDB} J.}, 2015.

\bibitem[\protect\citeauthoryear{Geng \bgroup \em et al.\egroup
  }{2021}]{OntoZSL}
Yuxia Geng, Jiaoyan Chen, Zhuo Chen, Jeff~Z. Pan, Zhiquan Ye, Zonggang Yuan,
  Yantao Jia, and Huajun Chen.
\newblock Ontozsl: Ontology-enhanced zero-shot learning.
\newblock In {\em {WWW}}, 2021.

\bibitem[\protect\citeauthoryear{Glimm \bgroup \em et al.\egroup
  }{2014}]{Hermit}
Birte Glimm, Ian Horrocks, Boris Motik, Giorgos Stoilos, and Zhe Wang.
\newblock Hermit: An {OWL} 2 reasoner.
\newblock {\em J. Autom. Reason.}, 2014.

\bibitem[\protect\citeauthoryear{Guo \bgroup \em et al.\egroup }{2016}]{KALE}
Shu Guo, Quan Wang, Lihong Wang, Bin Wang, and Li~Guo.
\newblock Jointly embedding knowledge graphs and logical rules.
\newblock In {\em {EMNLP}}, 2016.

\bibitem[\protect\citeauthoryear{Guo \bgroup \em et al.\egroup }{2018}]{RUGE}
Shu Guo, Quan Wang, Lihong Wang, Bin Wang, and Li~Guo.
\newblock Knowledge graph embedding with iterative guidance from soft rules.
\newblock In {\em {AAAI}}, 2018.

\bibitem[\protect\citeauthoryear{Guo \bgroup \em et al.\egroup }{2020}]{SLRE}
Shu Guo, Lin Li, Zhen Hui, Lingshuai Meng, Bingnan Ma, Wei Liu, Lihong Wang,
  Haibin Zhai, and Hong Zhang.
\newblock Knowledge graph embedding preserving soft logical regularity.
\newblock In {\em {CIKM}}, 2020.

\bibitem[\protect\citeauthoryear{Guu \bgroup \em et al.\egroup
  }{2015}]{traversing}
K.~Guu, J.~Miller, and P.~Liang.
\newblock Traversing knowledge graphs in vector space.
\newblock In {\em {EMNLP}}, 2015.

\bibitem[\protect\citeauthoryear{Hamilton \bgroup \em et al.\egroup
  }{2018}]{GQE}
W.~L. Hamilton, P.~Bajaj, M.~Zitnik, D.~Jurafsky, and JJ.ure Leskovec.
\newblock Embedding logical queries on knowledge graphs.
\newblock In {\em NeurIPS}, 2018.

\bibitem[\protect\citeauthoryear{Ho \bgroup \em et al.\egroup }{2018}]{RuLES}
Vinh~Thinh Ho, Daria Stepanova, Mohamed~H. Gad{-}Elrab, Evgeny Kharlamov, and
  Gerhard Weikum.
\newblock Rule learning from knowledge graphs guided by embedding models.
\newblock In {\em {ISWC}}, 2018.

\bibitem[\protect\citeauthoryear{Horrocks \bgroup \em et al.\egroup }{}]{HKS06}
Ian Horrocks, Oliver Kutz, and Ulrike Sattler.
\newblock The even more irresistible sroiq.
\newblock In {\em KR2006}.

\bibitem[\protect\citeauthoryear{Jain \bgroup \em et al.\egroup
  }{2018}]{TypeKGE}
Prachi Jain, Pankaj Kumar, Mausam, and Soumen Chakrabarti.
\newblock Type-sensitive knowledge base inference without explicit type
  supervision.
\newblock In {\em {ACL}}, 2018.

\bibitem[\protect\citeauthoryear{Jain \bgroup \em et al.\egroup
  }{2021}]{JTGS2021}
N.~Jain, T.~Tran, M.~H. Gad{-}Elrab, and D.~Stepanova.
\newblock Improving knowledge graph embeddings with ontological reasoning.
\newblock In {\em ISWC}, 2021.

\bibitem[\protect\citeauthoryear{Kotnis \bgroup \em et al.\egroup
  }{2021}]{BiQE}
B.~Kotnis, C.~Lawrence, and M.~Niepert.
\newblock Ans. complex queries in {KGs} with bidirectional sequence encoders.
\newblock In {\em {AAAI}}, 2021.

\bibitem[\protect\citeauthoryear{Krötzsch \bgroup \em et al.\egroup
  }{2008}]{KRH08}
Markus Krötzsch, S.~Rudolph, and P.~Hitzler.
\newblock Description logic rules.
\newblock 2008.

\bibitem[\protect\citeauthoryear{Li \bgroup \em et al.\egroup
  }{2020}]{SWJ-survey}
Weizhuo Li, Guilin Qi, and Qiu Ji.
\newblock Hybrid reasoning in knowledge graphs: Combing symbolic reasoning and
  statistical reasoning.
\newblock {\em Semantic Web}, 2020.

\bibitem[\protect\citeauthoryear{Lin \bgroup \em et al.\egroup
  }{2015}]{PTransE}
Y.~Lin, Z.~Liu, H.~Luan, M.~Sun, S.~Rao, and S.~Liu.
\newblock Modeling relation paths for representation learning of knowledge
  bases.
\newblock In {\em {EMNLP}}, 2015.

\bibitem[\protect\citeauthoryear{Liu \bgroup \em et al.\egroup
  }{2021}]{NewLook}
Lihui Liu, Boxin Du, Heng Ji, ChengXiang Zhai, and Hanghang Tong.
\newblock Neural-answering logical queries on knowledge graphs.
\newblock In {\em {KDD}}, 2021.

\bibitem[\protect\citeauthoryear{Meilicke \bgroup \em et al.\egroup
  }{2019}]{AnyBURL}
C.~Meilicke, M.~Chekol, D.~Ruffinelli, and H.~Stuckenschmidt.
\newblock Anytime bottom-up rule learning for knowledge graph completion.
\newblock In {\em {IJCAI}}, 2019.

\bibitem[\protect\citeauthoryear{Minervini \bgroup \em et al.\egroup
  }{2017}]{KGE_R}
P.~Minervini, L.~Costabello, E., V., and P.~Vandenbussche.
\newblock Regularizing knowledge graph embeddings via equivalence and inversion
  axioms.
\newblock In {\em {ECML/PKDD}}, 2017.

\bibitem[\protect\citeauthoryear{Minervini \bgroup \em et al.\egroup
  }{2020a}]{GNTP}
Pasquale Minervini, Matko Bosnjak, Tim Rockt{\"{a}}schel, Sebastian Riedel, and
  Edward Grefenstette.
\newblock Differentiable reasoning on large knowledge bases and natural
  language.
\newblock In {\em {AAAI}}, 2020.

\bibitem[\protect\citeauthoryear{Minervini \bgroup \em et al.\egroup
  }{2020b}]{CTP}
Pasquale Minervini, Sebastian Riedel, Pontus Stenetorp, Edward Grefenstette,
  and Tim Rockt{\"{a}}schel.
\newblock Learning reasoning strategies in end-to-end differentiable proving.
\newblock In {\em {ICML}}, 2020.

\bibitem[\protect\citeauthoryear{Nenov \bgroup \em et al.\egroup
  }{2015}]{RDFox}
Yavor Nenov, Robert Piro, Boris Motik, Ian Horrocks, Zhe Wu, and Jay Banerjee.
\newblock Rdfox: {A} highly-scalable {RDF} store.
\newblock In {\em {ISWC}}, 2015.

\bibitem[\protect\citeauthoryear{Nickel \bgroup \em et al.\egroup
  }{2011}]{RESCAL}
Maximilian Nickel, Volker Tresp, and Hans{-}Peter Kriegel.
\newblock A three-way model for collective learning on multi-relational data.
\newblock In {\em {ICML}}, 2011.

\bibitem[\protect\citeauthoryear{Niu \bgroup \em et al.\egroup }{2020}]{RPJE}
Guanglin Niu, Yongfei Zhang, Bo~Li, Peng Cui, Si~Liu, Jingyang Li, and Xiaowei
  Zhang.
\newblock Rule-guided compositional representation learning on knowledge
  graphs.
\newblock In {\em {AAAI}}, 2020.

\bibitem[\protect\citeauthoryear{Omran \bgroup \em et al.\egroup
  }{2018}]{RLvLR}
Pouya~Ghiasnezhad Omran, Kewen Wang, and Zhe Wang.
\newblock Scalable rule learning via learning representation.
\newblock In {\em {IJCAI}}, 2018.

\bibitem[\protect\citeauthoryear{Pan and Horrocks}{2006}]{PaHo06}
Jeff~Z. Pan and Ian Horrocks.
\newblock {OWL-Eu: Adding Customised Datatypes into OWL}.
\newblock {\em Journal of Web Semantics}, 2006.

\bibitem[\protect\citeauthoryear{Ren and Leskovec}{2020}]{BetaE}
Hongyu Ren and Jure Leskovec.
\newblock Beta embeddings for multi-hop logical reasoning in knowledge graphs.
\newblock In {\em NeurIPS}, 2020.

\bibitem[\protect\citeauthoryear{Ren \bgroup \em et al.\egroup
  }{2020}]{Query2box}
Hongyu Ren, Weihua Hu, and Jure Leskovec.
\newblock Query2box: Reasoning over knowledge graphs in vector space using box
  embeddings.
\newblock In {\em {ICLR}}, 2020.

\bibitem[\protect\citeauthoryear{Rockt{\"{a}}schel and Riedel}{2017}]{NTP}
T.~Rockt{\"{a}}schel and S.~Riedel.
\newblock End-to-end differentiable proving.
\newblock In {\em {NIPS}}, 2017.

\bibitem[\protect\citeauthoryear{Rosso \bgroup \em et al.\egroup }{2021}]{RETA}
Paolo Rosso, Dingqi Yang, Natalia Ostapuk, and Philippe Cudr{\'{e}}{-}Mauroux.
\newblock {RETA:} {A} schema-aware, end-to-end solution for instance completion
  in knowledge graphs.
\newblock In {\em {WWW}}, 2021.

\bibitem[\protect\citeauthoryear{Sadeghian \bgroup \em et al.\egroup
  }{2019}]{DRUM}
A.~Sadeghian, M.~Armandpour, P.~Ding, and D.~Wang.
\newblock {DRUM:} end-to-end differentiable rule mining on knowledge graphs.
\newblock In {\em NeurIPS}, 2019.

\bibitem[\protect\citeauthoryear{Song \bgroup \em et al.\egroup
  }{2021}]{Rot-Pro}
Tengwei Song, Jie Luo, and Lei Huang.
\newblock Rot-pro: Modeling transitivity by projection in knowledge graph
  embedding.
\newblock {\em NeurIPS}, 2021.

\bibitem[\protect\citeauthoryear{Sun \bgroup \em et al.\egroup }{2019}]{RotatE}
Zhiqing Sun, Zhi{-}Hong Deng, Jian{-}Yun Nie, and Jian Tang.
\newblock Rotate: Knowledge graph embedding by relational rotation in complex
  space.
\newblock In {\em {ICLR}}, 2019.

\bibitem[\protect\citeauthoryear{Trouillon \bgroup \em et al.\egroup
  }{2016}]{ComplEx}
T.~Trouillon, J.~Welbl, S.~Riedel, E., and G.~Bouchard.
\newblock Complex embeddings for simple link prediction.
\newblock In {\em {ICML}}, 2016.

\bibitem[\protect\citeauthoryear{Wang and Cohen}{2016}]{ProPPR+MF}
William~Y. Wang and William~W. Cohen.
\newblock Learning first-order logic embeddings via matrix factorization.
\newblock In {\em {IJCAI}}, 2016.

\bibitem[\protect\citeauthoryear{Wang \bgroup \em et al.\egroup
  }{2013}]{ProPPR}
William~Yang Wang, Kathryn Mazaitis, and William~W. Cohen.
\newblock Programming with personalized pagerank: a locally groundable
  first-order probabilistic logic.
\newblock In {\em {CIKM}}, 2013.

\bibitem[\protect\citeauthoryear{Wang \bgroup \em et al.\egroup
  }{2015}]{DBLP:conf/ijcai/WangWG15}
Q.~Wang, B.~Wang, and L.~Guo.
\newblock Knowledge base completion using embeddings and rules.
\newblock In {\em {IJCAI}}, 2015.

\bibitem[\protect\citeauthoryear{Wang \bgroup \em et al.\egroup }{2018a}]{LFDS}
G.~Wang, W.~Zhang, R.~Wang, Y.~Zhou, X.~Chen, W.~Zhang, H.~Zhu, and H.~Chen.
\newblock Label-free distant supervision for relation extraction via knowledge
  graph embedding.
\newblock In {\em {EMNLP}}, 2018.

\bibitem[\protect\citeauthoryear{Wang \bgroup \em et al.\egroup }{2018b}]{TARE}
M.~Wang, E.~Rong, H.~Zhuo, and H.~Zhu.
\newblock Embedding knowledge graphs based on transitivity and asymmetry of
  rules.
\newblock In {\em {PAKDD}}, 2018.

\bibitem[\protect\citeauthoryear{Wang \bgroup \em et al.\egroup
  }{2020}]{Neural-Num-LP}
Po{-}Wei Wang, Daria Stepanova, Csaba Domokos, and J.~Zico Kolter.
\newblock Differentiable learning of numerical rules in knowledge graphs.
\newblock In {\em {ICLR}}, 2020.

\bibitem[\protect\citeauthoryear{Wang \bgroup \em et al.\egroup }{2021}]{TAGAT}
Yuzhuo Wang, Hongzhi Wang, Junwei He, Wenbo Lu, and Shuolin Gao.
\newblock {TAGAT:} type-aware graph attention networks for reasoning over
  knowledge graphs.
\newblock {\em Knowl. Based Syst.}, 233:107500, 2021.

\bibitem[\protect\citeauthoryear{Wei \bgroup \em et al.\egroup }{2015}]{INS-ES}
Zhuoyu Wei, Jun Zhao, Kang Liu, Zhenyu Qi, Zhengya Sun, and Guanhua Tian.
\newblock Large-scale knowledge base completion: Inferring via grounding
  network sampling over selected instances.
\newblock In {\em {CIKM}}, 2015.

\bibitem[\protect\citeauthoryear{Wiharja \bgroup \em et al.\egroup
  }{2020}]{SIC}
Kemas Wiharja, Jeff~Z. Pan, Martin~J. Kollingbaum, and Yu~Deng.
\newblock {Schema Aware Iterative Knowledge Graph Completion}.
\newblock {\em JoWS}, 2020.

\bibitem[\protect\citeauthoryear{Wong \bgroup \em et al.\egroup
  }{2021}]{KGPretrain4RS}
C.~Wong, F.~Feng, W.~Zhang, C.~Vong, H.~Chen, Y.~Zhang, P.~He, H.~Chen,
  K.~Zhao, and H.~Chen.
\newblock Improving conversational recommender system by pretraining
  billion-scale knowledge graph.
\newblock In {\em {ICDE}}, 2021.

\bibitem[\protect\citeauthoryear{Wu \bgroup \em et al.\egroup }{2015}]{SePLi}
Fei Wu, Jun Song, Yi~Yang, Xi~Li, Zhongfei~(Mark) Zhang, and Yueting Zhuang.
\newblock Structured embedding via pairwise relations and long-range
  interactions in knowledge base.
\newblock In {\em {AAAI}}, 2015.

\bibitem[\protect\citeauthoryear{Xie \bgroup \em et al.\egroup }{2016}]{TKRL}
Ruobing Xie, Zhiyuan Liu, and Maosong Sun.
\newblock Representation learning of knowledge graphs with hierarchical types.
\newblock In {\em {IJCAI}}, 2016.

\bibitem[\protect\citeauthoryear{Yang \bgroup \em et al.\egroup
  }{2015}]{DistMult}
B.~Yang, W.~Yih, X.~He, J.~Gao, and L.~Deng.
\newblock Embedding entities and relations for learning and inference in
  knowledge bases.
\newblock In {\em {ICLR}}, 2015.

\bibitem[\protect\citeauthoryear{Yang \bgroup \em et al.\egroup
  }{2017}]{NeuralLP}
Fan Yang, Zhilin Yang, and William~W. Cohen.
\newblock Differentiable learning of logical rules for knowledge base
  reasoning.
\newblock In {\em {NIPS}}, 2017.

\bibitem[\protect\citeauthoryear{Zhang \bgroup \em et al.\egroup }{2018}]{HRS}
Zhao Zhang, Fuzhen Zhuang, Meng Qu, Fen Lin, and Qing He.
\newblock Knowledge graph embedding with hierarchical relation structure.
\newblock In {\em {EMNLP}}, 2018.

\bibitem[\protect\citeauthoryear{Zhang \bgroup \em et al.\egroup
  }{2019}]{IterE}
W.~Zhang, B.~Paudel, L.~Wang, J.~Chen, H.~Zhu, W.~Zhang, A.~Bernstein, and
  H.~Chen.
\newblock Iteratively learning embeddings and rules for knowledge graph
  reasoning.
\newblock In {\em {WWW}}, 2019.

\bibitem[\protect\citeauthoryear{Zhang \bgroup \em et al.\egroup
  }{2020a}]{TransRHS}
Fuxiang Zhang, Xin Wang, Zhao Li, and Jianxin Li.
\newblock Transrhs: {A} representation learning method for knowledge graphs
  with relation hierarchical structure.
\newblock In {\em {IJCAI}}, 2020.

\bibitem[\protect\citeauthoryear{Zhang \bgroup \em et al.\egroup
  }{2020b}]{HAKE}
Zhanqiu Zhang, Jianyu Cai, Yongdong Zhang, and Jie Wang.
\newblock Learning hierarchy-aware knowledge graph embeddings for link
  prediction.
\newblock In {\em {AAAI}}, 2020.

\bibitem[\protect\citeauthoryear{Zhang \bgroup \em et al.\egroup
  }{2021a}]{AIOpen-survey}
Jing Zhang, Bo~Chen, Lingxi Zhang, Xirui Ke, and Haipeng Ding.
\newblock Neural, symbolic and neural-symbolic reasoning on knowledge graphs.
\newblock {\em AI Open}, 2021.

\bibitem[\protect\citeauthoryear{Zhang \bgroup \em et al.\egroup
  }{2021b}]{PKGM}
W.~Zhang, C.~Man Wong, G.~Ye, B.~Wen, W.~Zhang, and H.~Chen.
\newblock Billion-scale pre-trained e-commerce product knowledge graph model.
\newblock In {\em {ICDE}}, 2021.

\bibitem[\protect\citeauthoryear{Zhang \bgroup \em et al.\egroup
  }{2021c}]{ConE}
Z.~Zhang, J.~Wang, J.~Chen, S.~Ji, and F.~Wu.
\newblock Cone: Cone embeddings for multi-hop reasoning over knowledge graphs.
\newblock In {\em NeurIPS}, 2021.

\bibitem[\protect\citeauthoryear{Zhang}{2017}]{dORC}
Wen Zhang.
\newblock Knowledge graph embedding with diversity of structures.
\newblock In {\em {WWW}}, 2017.

\end{thebibliography}

\end{document}